\documentclass[11pt]{article}

\usepackage[preprint]{acl}

\usepackage{times}
\usepackage{latexsym}
\usepackage{amsmath,amsfonts}
\usepackage{array}
\usepackage{booktabs}
\usepackage{multirow}

\usepackage[T1]{fontenc}

\usepackage[utf8]{inputenc}

\usepackage{microtype}

\usepackage{inconsolata}

\usepackage{graphicx}
\usepackage[table]{xcolor}

\usepackage{hyperref}
\usepackage{url}
\usepackage{xspace}
\newif\ifwidepage
  \widepagetrue
  \ifwidepage
  \fi

\definecolor{TitleBlue}{RGB}{24,116,205}
\definecolor{TitleOrange}{RGB}{255,127,0}
\definecolor{tableavg}{RGB}{228,234,242}
\definecolor{tablegray}{RGB}{246,247,249}
\definecolor{tablemodel}{RGB}{238,244,251}
\definecolor{tablevanilla}{RGB}{242,244,247}
\definecolor{tablevanillahead}{RGB}{226,231,238}
\definecolor{tableours}{RGB}{232,242,255}
\definecolor{tableourshead}{RGB}{199,222,248}

\newcommand{\AQ}{\textit{Answer$\rightarrow$Query}\xspace}
\newcommand{\IIflow}{\textit{Image$\rightarrow$Image}\xspace}
\newcommand{\AIflow}{\textit{Answer$\rightarrow$Image}\xspace}
\newcommand{\QIflow}{\textit{Query$\rightarrow$Image}\xspace}
\newcommand{\QQflow}{\textit{Query$\rightarrow$Query}\xspace}
\newcommand{\VRWWidth}{\textit{VRW-Width}\xspace}
\newcommand{\VRWEnd}{\textit{VRW-End}\xspace}
\newcommand{\AnchorStrength}{\textit{AnchorStrength}\xspace}
\newcommand{\score}[2]{#1\if\relax\detokenize{#2}\relax\else {\color{black!55}\scriptsize(#2)}\fi}
\newcommand{\bestscore}[2]{\textbf{#1}\if\relax\detokenize{#2}\relax\else {\color{black!55}\scriptsize(#2)}\fi}
\newcommand{\secondscore}[2]{\underline{#1}\if\relax\detokenize{#2}\relax\else {\color{black!55}\scriptsize(#2)}\fi}

\title{%
  \textcolor{TitleBlue!100!TitleOrange}{T}\textcolor{TitleBlue!93!TitleOrange}{h}\textcolor{TitleBlue!86!TitleOrange}{e} %
  \textcolor{TitleBlue!79!TitleOrange}{E}\textcolor{TitleBlue!72!TitleOrange}{b}\textcolor{TitleBlue!64!TitleOrange}{b} %
  \textcolor{TitleBlue!57!TitleOrange}{a}\textcolor{TitleBlue!50!TitleOrange}{n}\textcolor{TitleBlue!43!TitleOrange}{d} %
  \textcolor{TitleBlue!36!TitleOrange}{F}\textcolor{TitleBlue!29!TitleOrange}{l}\textcolor{TitleBlue!21!TitleOrange}{o}\textcolor{TitleBlue!14!TitleOrange}{w} %
  \textcolor{TitleBlue!7!TitleOrange}{o}\textcolor{TitleBlue!0!TitleOrange}{f} %
  \textcolor{TitleBlue!0!TitleOrange}{M}\textcolor{TitleBlue!7!TitleOrange}{u}\textcolor{TitleBlue!14!TitleOrange}{l}\textcolor{TitleBlue!21!TitleOrange}{t}\textcolor{TitleBlue!29!TitleOrange}{i}\textcolor{TitleBlue!36!TitleOrange}{m}\textcolor{TitleBlue!43!TitleOrange}{o}\textcolor{TitleBlue!50!TitleOrange}{d}\textcolor{TitleBlue!57!TitleOrange}{a}\textcolor{TitleBlue!64!TitleOrange}{l} %
  \textcolor{TitleBlue!71!TitleOrange}{F}\textcolor{TitleBlue!79!TitleOrange}{o}\textcolor{TitleBlue!86!TitleOrange}{c}\textcolor{TitleBlue!93!TitleOrange}{u}\textcolor{TitleBlue!100!TitleOrange}{s}:

  Scheduling Visual Relay Windows for Grounded VLM Reasoning%
}

\author{
  \textbf{Wencheng Ye\textsuperscript{1}},
  \textbf{Yi Bin\textsuperscript{1}\thanks{Corresponding author.}},
  \textbf{Yujuan Ding\textsuperscript{2}},
  \textbf{Hongye Fang\textsuperscript{1}},
  \textbf{Zheng Wang\textsuperscript{1}},
  \textbf{Xing Xu\textsuperscript{1}}, \\
  \textbf{Jingkuan Song\textsuperscript{1}},
  \textbf{Yun Zhang\textsuperscript{3}},
  \textbf{Sirui Da\textsuperscript{1}},
  \textbf{Heng Tao Shen\textsuperscript{1}} \\
  \textsuperscript{1}School of Computer Science and Technology, Tongji University \\
  \textsuperscript{2}Department of Computing, The Hong Kong Polytechnic University \\
  \textsuperscript{3}School of Artificial Intelligence, Shanghai Jiao Tong University
}

\begin{document}
\maketitle
\begin{abstract}
Vision-language models increasingly succeed on multimodal reasoning benchmarks, yet their visual evidence often becomes unstable once it enters the language stack, weakening evidence-grounded reasoning. To understand this fragility, we examine the internal dynamics of VLMs through a mechanistic lens and uncover a stable three-stage redistribution of multimodal attention focus across depth: an early question-conditioned organization, a critical middle visual-dominant relay, and a late return to answer formation. We operationalize the middle phase as the Visual Relay Window (VRW), and show that its geometry varies with task demand, is causally tied to grounded generation, and distinguishes unsupported answers from stronger reasoning trajectories. Guided by this internal rhythm, we propose TRACE, a task-adaptive inference-time control framework with lightweight trained modules. It reshapes relay allocation during prefill and preserves assembled visual support after handoff during decoding. Across four open-weight VLM backbones and seven benchmarks, TRACE delivers large gains on grounding-sensitive settings, improving them by 4.33 points on average and by up to 6.6 points, while also improving reasoning-heavy tasks. These results show that explicitly controlling multimodal focus across depth offers a unified and effective mechanism for strengthening evidence-grounded multimodal reasoning. The code is available at:
\url{https://github.com/gooogleshanghai/visual-relay-window}.
\end{abstract}

\noindent\textbf{Keywords:} Vision-language models, multimodal reasoning, hallucination mitigation.

\section{Introduction}
\begin{figure*}[t]
\centering
\includegraphics[width=1.0\textwidth]{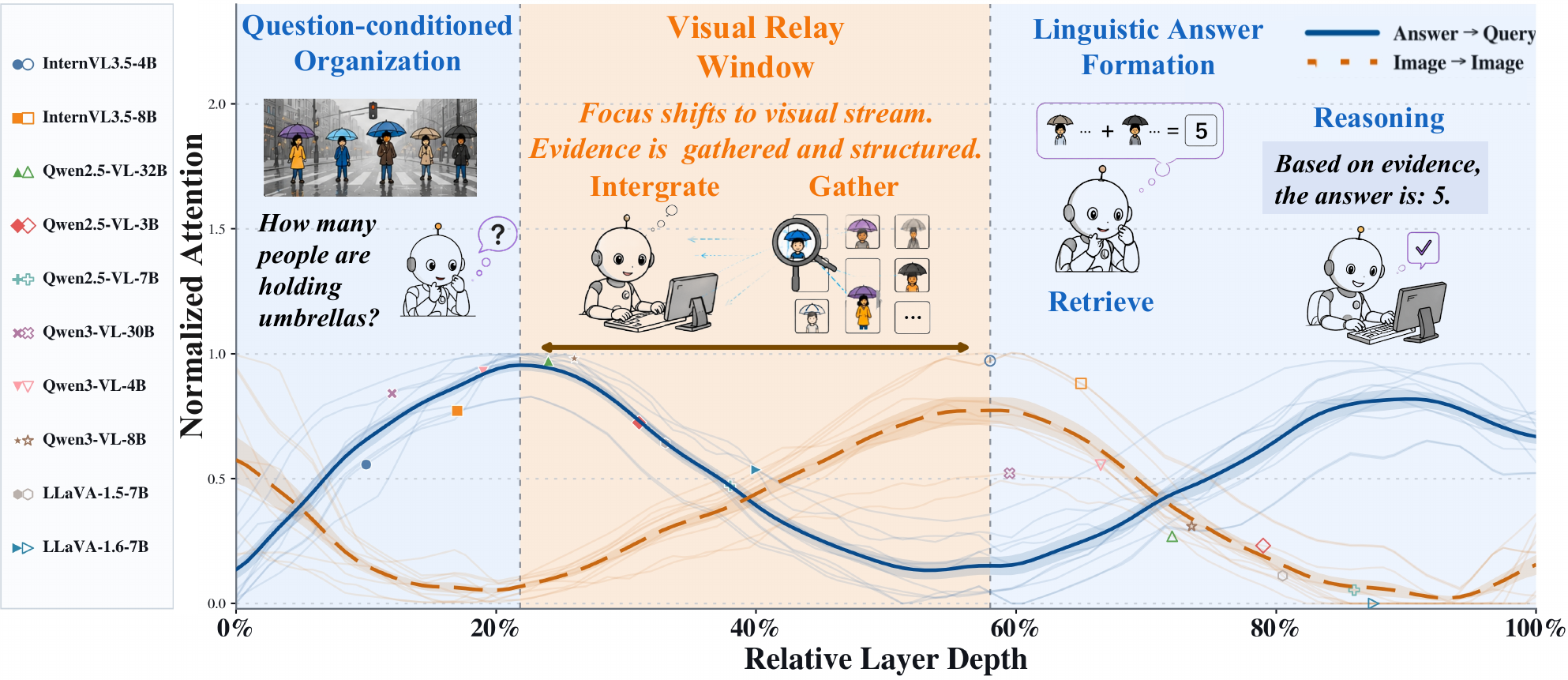}
\caption{A unified three-stage relay pattern across vision-language models. Across ten open VLMs, the jointly normalized
  Answer→Query and Image→Image trajectories show a consistent depth-wise redistribution of multimodal focus: early layers
  emphasize question-conditioned response organization, middle layers allocate computation to visual consolidation, and
  later layers shift back toward answer formation. The horizontal bracket marks the relay interval, while paired markers
  match individual models to their trajectories.}
\label{fig:cross_model_vrw}
\end{figure*}
Vision-language models (VLMs) have advanced rapidly in both scale and capability, with recent systems showing  competitive reasoning ability and strong performance on broad multimodal benchmarks~\cite{zhang2024vision,yang2025qwen3,wang2025internvl3,yue2024mmmu,lu2024mathvista,thawakar2025llamav}. Yet this progress also sharpens a central requirement: visual evidence must remain available across multiple inference steps and be reorganized into answer-relevant structure~\cite{li2025lost,liu2024survey}. This is difficult because once visual information is absorbed into the language stack, its influence can attenuate with depth, become entangled with language organization, or drift as generation unfolds~\cite{zhong2024investigating,xu2026more}.

Recent methods improve visual evidence support in VLMs through diverse control handles anchored to different  actionable signals: contrastive decoding exploits the sensitivity of generation to visual perturbation~\cite{leng2024mitigating,zhao2024mitigating}, attention reallocation targets insufficient attention paid to visual tokens~\cite{liu2024paying,zhang2026prefill}, and activation steering targets activation patterns associated with ungrounded generation~\cite{wu2025sharp,su2025activation,liu2025reducing,yin2026dynamic}. Together, these advances show that failures of visual evidence support can often be mitigated at inference time once the right signal is acted on at the right place. What remains less clear is how these signals, each anchored to a different point of computation, relate to one another over the course of inference: whether they should be treated as separate cues calling for separate fixes, or as partial snapshots of a single evolving process governing how visual evidence is organized and used. As a result, it remains  difficult to tell how intervention should track the changing role of visual evidence as reasoning unfolds.

Addressing this requires  a more explicit account of multimodal computation during inference. Prior mechanistic studies show that attention-based analysis can reveal functionally meaningful structure inside VLMs, identifying how visual information is redistributed across layers, token groups, and components~\cite{basu2024understanding,zhang2025cross,li2026causal,jiang2025devils,shanbhagprocess}. Through such analyses, they find that visually relevant transfer is concentrated in a subset of middle layers and fine-grained details are retrieved more locally from image tokens~\cite{zhang2025cross,kaduri2025s,jiang2025devils}. We therefore ask a more structural question: \textit{how does multimodal focus evolve over the course of inference as a VLM reasons toward an answer?}

To answer it, we track two attention probes across layers: an answer-side probe measuring how strongly answer tokens look back to the question, and a visual-side probe measuring how strongly visual tokens attend to one another. Aggregated attention mass under these two probes gives a compact proxy for how multimodal interaction is redistributed. What emerges is a clear  three-stage pattern, where the two probes behave like opposing currents over depth. In the early layers, attention remains dominated by question-conditioned organization: the model is still aligning the visual input with the query and establishing an answer-relevant frame, so the question-directed probe rises while intra-visual attention falls. Midway through the network, the balance shifts toward intra-visual consolidation and attention to the question recedes, where visual tokens interact more strongly with one another and answer-relevant evidence is actively assembled. In the later layers, computation returns to language-side answer formation, where the assembled support is handed off to generation. Figure~\ref{fig:cross_model_vrw} sketches this progression and shows that the same three-stage pattern appears consistently across models, marking a limited interval in which visual consolidation dominates before computation returns to language side.

Crucially, this relay geometry is not fixed. Our analysis suggests that it shifts with the evidence demands of the question: tasks that require broader visual aggregation such as counting tend to exhibit longer and later relay phases, whereas tasks that place greater weight on downstream reasoning tend to exhibit earlier handoff from visual consolidation to answer formation. Success therefore depends not only on whether relevant visual content is present, but also on whether the timing of relay match the computation the task requires. Mismatch can arise in either direction. If relay ends too early, answer-relevant evidence may remain only partially assembled before generation begins; if it persists too long, computation that would otherwise support later reasoning will be occupied by unneeded visual processing. This interpretation is further supported by causal analyses: perturbing states within the relay interval changes whether later responses remain grounded far more strongly than comparable perturbations outside it. Stronger Thinking variants likewise exhibit more stable relay matching and stronger post-handoff anchoring, suggesting that part of their advantage may lie in allocating this intermediate stage more appropriately.

These findings make relay geometry a natural target for intervention. We therefore propose TRACE (\textbf{T}ask-adaptive \textbf{R}elay \textbf{A}nchoring and \textbf{C}ontrolled \textbf{E}vidence Scheduling, an inference-time control framework with lightweight trained modules that regulates how visual support is assembled before handoff and retained afterward. TRACE first uses lightweight prefill-time attention statistics to predict which layers are likely to participate in relay, then adaptively expands or contracts relay allocation through task-aware scheduling, and finally anchors the selected visual support after handoff so that relevant evidence remains available during decoding. 

We evaluate TRACE on four open-weight VLM backbones from two families across seven benchmarks spanning document understanding, visual reasoning, and hallucination-sensitive grounding. It delivers large gains on grounding-sensitive settings, with an average improvement of 4.33 points and gains of up to 6.6 points, while also improving reasoning-heavy settings consistently. The intervention also produces the intended internal shifts: grounding-sensitive benchmarks  benefit from broader, later relay phases, whereas reasoning-heavy cases improve when relay becomes shorter and hands off earlier. This highlights the unified feature of our approach: rather than designing separate fixes for hallucination mitigation or logical reasoning, we cast the diverse demands of downstream tasks into a single underlying mechanism.

Our contributions are threefold:
\begin{itemize}
    \item We identify a stable stage-wise redistribution of multimodal focus across depth and operationalize its middle relay phase as the \textit{Visual Relay Window} (VRW), establishing a structural measure for internal evidence assembly.
    
    \item  We demonstrate that this geometry dynamically adapts to specific task demands, and establish a causal link between optimal relay timing and grounded generation, offering a mechanistic explanation for typical hallucinations and capability differences across models.
    
        \item We introduce TRACE, an intervention framework that reshapes relay allocation during prefill and preserves assembled support during decoding. By viewing the diverse demands of downstream tasks through a single computational lens, it addresses both hallucination mitigation and complex reasoning under a unified control principle.
\end{itemize}
\section{Related Work}

\subsection{Visual Evidence Degradation in Multimodal Reasoning}
Vision-language models (VLMs) have increasingly shifted from perception-oriented tasks toward multi-step, evidence-dependent reasoning\cite{zhang2024vision,thawakar2025llamav,zhou2026visualizing}. This transition is especially visible on benchmarks such as MMMU\cite{yue2024mmmu} and MathVista\cite{lu2024mathvista}, where success depends on carrying visual evidence across multiple inference steps, and is widely tied to a persistent modality gap between visual and language representations\cite{bai2024hallucination,liu2024survey}. Since visual tokens are projected into the language stack through a lightweight connector, they often behave as weakly aligned soft prompts whose influence degrades with depth and generation length\cite{li2025lost,chen2026babyvision}. Modern VLMs therefore frequently fall back to language priors over visual evidence, leading to visual neglect across perception, spatial understanding, and long-form reasoning settings\cite{zhao2024difficult,chandhok2025response,alam2026spatial,xi2026large,xu2026more,zhong2024investigating}. These findings shift attention from visual access alone to how visual evidence is maintained through reasoning.

\subsection{Mechanistic Analysis of Multimodal Computation}
To understand why generation becomes less visually grounded during reasoning, recent work has increasingly used attention patterns and token-group interactions as informative probes of multimodal computation inside VLMs~\cite{shanbhagprocess,zhang2025cross,saporita2026fg}.  Through these analyses, several studies show that intermediate layers play an especially important role
  in consolidating visual evidence and aligning it with language
  representations~\cite{yang2026circuit,kaduri2025s,jiang2025devils}. Causal tracing and circuit-level
  analyses further suggest that object representations and image-to-text transfer pathways are often
  concentrated in a subset of layers and
  components~\cite{li2026causal,zhang2025redundancy,basu2024understanding}. Several works also report
  failure sources such as representational interference, attention sinks, and visual information
  dilution~\cite{kang2025see,chenvocabulary,savietto2026geometry,zhang2024seeing}. These findings together  suggest
  that visual information is redistributed non-uniformly across depth, but the stage-wise structure that
  governs evidence buildup remains less explicitly characterized. Using attention as a lens on how
  interaction is redistributed across token groups and depth, our analysis reveals a relay pattern in
  which visual consolidation and language-side organization exhibit a clear stage-wise trade-off, and
  operationalizes this structure as the Visual Relay Window.

\subsection{Inference-Time Control for Vision-Language Models}
A large body of previous work attempts to mitigate failures of visual evidence support through inference-time intervention. One prominent direction is contrastive decoding, including VCD\cite{leng2024mitigating}, MARINE\cite{zhao2024mitigating}, and several perturbation-based variants\cite{park2025second,jiang2026kvsmooth}, which contrast original and perturbed visual representations to suppress unsupported generations.  These approaches mainly operate at the final decoding stage, therefore cannot directly repair upstream evidence formation errors. Another line of work intervenes in latent representations through activation steering or hidden-state editing\cite{wu2025sharp,wang2025adaptive,su2025activation,liu2025reducing}, which may alter entangled reasoning-relevant representations\cite{sivakumar2025steervlm,cheng2026rfi,yin2026dynamic}. More recently, attention-level and prefill-stage interventions have emerged as a promising alternative. Methods such as PTI\cite{zhang2026prefill}, PAI\cite{liu2024paying}, and AVAM\cite{zeng2025avam} show that earlier intervention points can improve visual evidence support by reallocating attention or preserving visual context before decoding begins\cite{zhang2026vib,fu2026diagnosing}. Our work builds on these advances by explicitly modeling relay geometry and using it as a unified control target. This intervention improves grounded reasoning directly, while hallucination reduction emerges from the same mechanism.

\section{Structure-Aware Control in VLMs}
\label{sec:vrw}
We first identify a relay structure that organizes multimodal evidence flow and establish it as a measurable target for grounded generation. We then show how TRACE operationalizes this structure as an inference-time control mechanism.

\subsection{Attention Probes for Depth-Wise Multimodal Focus}
\label{sec:relay_probes}

Our goal is to examine how multimodal focus evolves across depth as a VLM reasons. To do so, we track a small set of token-group attention flows that capture complementary aspects of this process. One reflects question-conditioned response organization, while the other reflects consolidation within the visual stream. Consider a vision-language model that takes an image-question pair $(I,q)$ as input and autoregressively generates an answer sequence $y=(y_1,\ldots,y_T)$. Let the model contain $L$ Transformer layers, and let $\mathbf{A}^{(l)} \in \mathbb{R}^{N \times N}$ denote the self-attention matrix at layer $l$. We partition the sequence into three groups,
\begin{equation}
\mathcal{V} \cup \mathcal{Q} \cup \mathcal{Y} = \{1,\ldots,N\},
\end{equation}
where $\mathcal{V}$, $\mathcal{Q}$, and $\mathcal{Y}$ respectively denote visual, question, and answer tokens. We first define an answer-side probe that measures how strongly the model refers back to the question while forming a response:
\begin{equation}
M^{(l)}_{\mathrm{A}\rightarrow\mathrm{Q}}
=
\frac{1}{|\mathcal{Y}|}
\sum_{i \in \mathcal{Y}}
\sum_{j \in \mathcal{Q}}
\mathbf{A}^{(l)}_{ij},
\end{equation}
We next define a visual-side probe for intra-visual consolidation:
\begin{equation}
M^{(l)}_{\mathrm{I}\rightarrow\mathrm{I}}
=
\frac{1}{|\mathcal{V}|}
\sum_{i \in \mathcal{V}}
\sum_{j \in \mathcal{V}}
\mathbf{A}^{(l)}_{ij}.
\end{equation}

For autoregressive decoding, the answer-side probe aggregates the corresponding attention mass over answer-generation steps, yielding a layerwise estimate of question-conditioned response formation. The visual probe is computed within each layer as a summary of intra-visual consolidation. Together, these two probes provide a compact view of how multimodal computation is distributed between language-side organization and visual evidence assembly across depth.


\subsection{A Shared Relay Structure Across VLMs}
\label{sec:shared_relay}
Cognitive accounts of scene understanding describe a stage-wise progression, which moves from perceptual encoding to evidence integration and response selection~\cite{vanmaanen2021stages,epstein2019scene}. Consistent with this view, after aligning layers by relative depth and jointly normalizing the two relay channels, the balance between the two probes likewise reveals a three-stage trajectory: an early question-conditioned regime, a middle visual-dominant regime, and a late return to response formation. Figure~\ref{fig:cross_model_vrw} shows that this three-stage pattern survives substantial architectural variation, covering ten checkpoints from Qwen-VL\cite{qwen2025qwen25technicalreport,yang2025qwen3}, InternVL\cite{wang2025internvl3}, and LLaVA\cite{liu2024improved} on a balanced probing set of 2{,}000 image-question pairs spanning four task families. Despite large differences in model size, tokenizer layout, and absolute attention scale, the same phase ordering persists. What changes across models is the geometry of the middle visual-dominant segment, especially its extent and termination depth. We therefore isolate this segment and operationalize it as the \emph{Visual Relay Window} (VRW).

\begin{table}[t]
\centering
\caption{Top probe pairs ranked by reciprocal correlation on the probing split. Lower correlation indicates a stronger reciprocal trend.}
\label{tab:probe_pairs}
\scriptsize
\setlength{\tabcolsep}{2.6pt}
\renewcommand{\arraystretch}{1.05}
\begin{tabular}{lcc}
\toprule
Probe pair & Source relation & Corr. $\downarrow$ \\
\midrule
\AQ / \IIflow & Disjoint-source & \textbf{-0.93} \\
\AQ / \AIflow & Same-source & -0.87 \\
\QQflow / \IIflow & Disjoint-source & -0.84 \\
\AQ / \QIflow & Disjoint-source & -0.73 \\
\bottomrule
\end{tabular}
\end{table}
\subsection{Operationalizing the Visual Relay Window}
\label{sec:operational_vrw}

We formalize the VRW as an operationally estimated middle-layer segment in which visual evidence dominates before answer commitment. Among non-identical probe pairs, \AQ and \IIflow exhibit the clearest reciprocal depth-wise structure on the probing split (Table~\ref{tab:probe_pairs}), which makes them a suitable basis for this operational summary. To compare samples and model families, we jointly normalize the two relay channels within each sample, and define the relay-dominance score
\begin{equation}
R^{(l)}
=
\overline{M}^{(l)}_{\mathrm{I}\rightarrow\mathrm{I}}
-
\overline{M}^{(l)}_{\mathrm{A}\rightarrow\mathrm{Q}}.
\end{equation}
Higher values indicate stronger visual consolidation, while lower values indicate stronger question-conditioned response organization. Intuitively, $R^{(l)}$ is a signed summary of which side currently dominates the computation at layer $l$. In practice, before computing \(R^{(l)}\), we jointly normalize \(M^{(l)}_{\mathrm{I}\rightarrow\mathrm{I}}\) and \(M^{(l)}_{\mathrm{A}\rightarrow\mathrm{Q}}\) within each sample by min-max scaling them over layers to \([0,1]\).

To focus on the middle relay phase, we define the relative depth \(\rho_l = l/L\) and use the heuristic setting \(\rho_l \in [0.2,0.8]\) together with a tolerance margin \(\delta=0.1\). This restricts the estimate to the network interior and suppresses small local ripples. We then define the relay peak as
\begin{equation}
l^\star = \arg\max_{l:\,\rho_l \in [0.2,0.8]} R^{(l)}.
\end{equation}
The candidate interval \([s_0,e_0]\) is the maximal contiguous interval containing \(l^\star\) such that
\begin{equation}
R^{(l)} \ge R^{(l^\star)} - \delta,
\qquad
\forall l \in [s_0,e_0].
\end{equation}
This candidate interval permits local fluctuations while requiring the retained segment to remain close to the middle-layer relay peak. The final VRW is the longest contiguous sub-interval \([s,e] \subseteq [s_0,e_0]\) that still contains \(l^\star\) and preserves the global handoff trend:
\begin{equation}
\overline{M}^{(e)}_{\mathrm{I}\rightarrow\mathrm{I}} - \overline{M}^{(s)}_{\mathrm{I}\rightarrow\mathrm{I}} > 0,
\qquad
\overline{M}^{(e)}_{\mathrm{A}\rightarrow\mathrm{Q}} - \overline{M}^{(s)}_{\mathrm{A}\rightarrow\mathrm{Q}} < 0.
\end{equation}

From the final VRW boundaries, we derive two statistics used throughout the paper:
\begin{equation}
\textit{VRW-Width} = \frac{e-s+1}{L},
\qquad
\textit{VRW-End} = \frac{e}{L}.
\end{equation}
\VRWWidth measures the relative extent of the relay phase, while \VRWEnd measures the depth at which it terminates.

\subsection{Task-Dependent Relay Geometry}
\label{sec:task_geometry}

\begin{figure}[t]
\centering
\includegraphics[width=1.0\columnwidth]{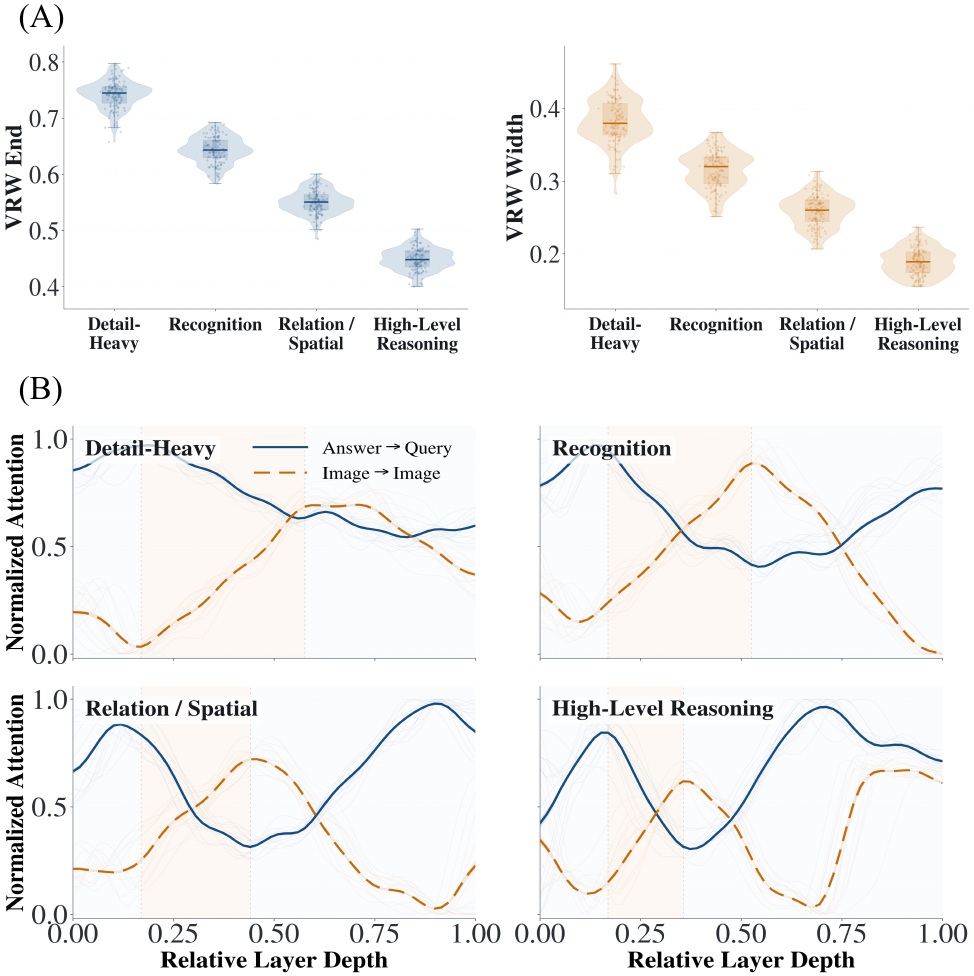}
\caption{Task-dependent geometry of the Visual Relay Window on Qwen3-VL-4B. (A) Distributions of \VRWEnd and \VRWWidth across task families. (B) Task-wise mean relay curves for four task families.}
\label{fig:task_vrw}
\end{figure}

The cross-model regularity above motivates an immediate question: is the relay window fixed or highly adaptive? To answer this, we group evaluation samples into four families—detail-heavy perception, recognition, relation/spatial reasoning, and high-level reasoning—and test whether relay geometry shifts systematically with evidence demand.

As Figure~\ref{fig:task_vrw} shows, relay geometry is strongly task-dependent (one-way ANOVA,  $p<0.005$). Detail-heavy tasks tend to produce broader windows and later relay termination, indicating that visual evidence continues to be assembled deeper into the stack. High-level reasoning tasks, by contrast, often exhibit shorter windows and earlier handoff, while recognition and relation-oriented tasks occupy intermediate regimes.  Taken together, these patterns show that the redistribution of multimodal focus is itself task-sensitive, which  reflects a control policy that the model may have acquired during massive training, allocating relay length and termination depth to match what each task's evidence-gathering process requires.

\subsection{Relay Mismatch as an Evidence Assembly Failure}
\label{sec:relay_mismatch}

Task-dependent relay geometry suggests a natural failure mode: the relay can be mismatched to the evidence demand of the question. Under this view, unsupported generation arises when the model begins linguistic commitment before the necessary visual structure has been fully consolidated. To probe this mismatch, we use HaloQuest\cite{wang2024haloquest} under repeated decoding with temperature \(1.0\), sampling 5 continuations for each of 200 image-question pairs. We classify each continuation according to the HaloQuest correctness criterion. To quantify whether answer formation remains tied to visual evidence after relay handoff, we measure answer-to-visual attention at each layer:
\begin{equation}
M^{(l)}_{\mathrm{A}\rightarrow\mathrm{V}}
=
\frac{1}{|\mathcal{Y}|}
\sum_{i \in \mathcal{Y}}
\sum_{j \in \mathcal{V}}
\mathbf{A}^{(l)}_{ij}.
\end{equation}
We then define \AnchorStrength as the average of this quantity after the relay window terminates:
\begin{equation}
\small
\textit{AnchorStrength}
=
\frac{1}{|L_{\mathrm{post}}|}
\sum_{l \in L_{\mathrm{post}}}
M^{(l)}_{\mathrm{A}\rightarrow\mathrm{V}},
\quad
L_{\mathrm{post}}
=
\{\, l \mid l > e \,\}.
\end{equation}
where $e$ denotes the VRW end layer.

\begin{figure}[t]
\centering
\includegraphics[width=1.0\columnwidth]{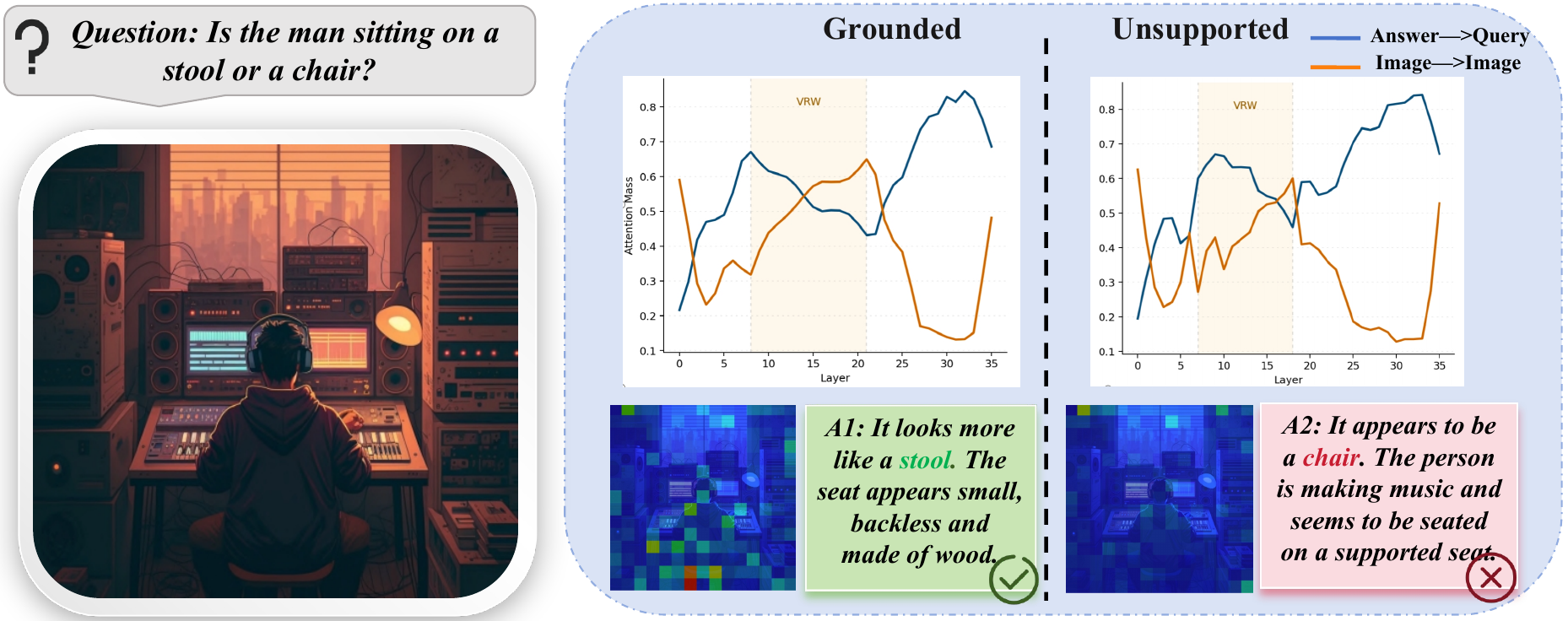}
\caption{Case study of relay bifurcation for the same image-question pair. The grounded trajectory sustains a broader relay phase and remains visually supported, whereas the unsupported trajectory exits earlier and drifts toward language-side priors. Lower overlays visualize answer-to-visual attention after relay termination.}
\label{fig:hallucination_case}
\end{figure}

\begin{figure}[t]
\centering
\includegraphics[width=1.0\columnwidth]{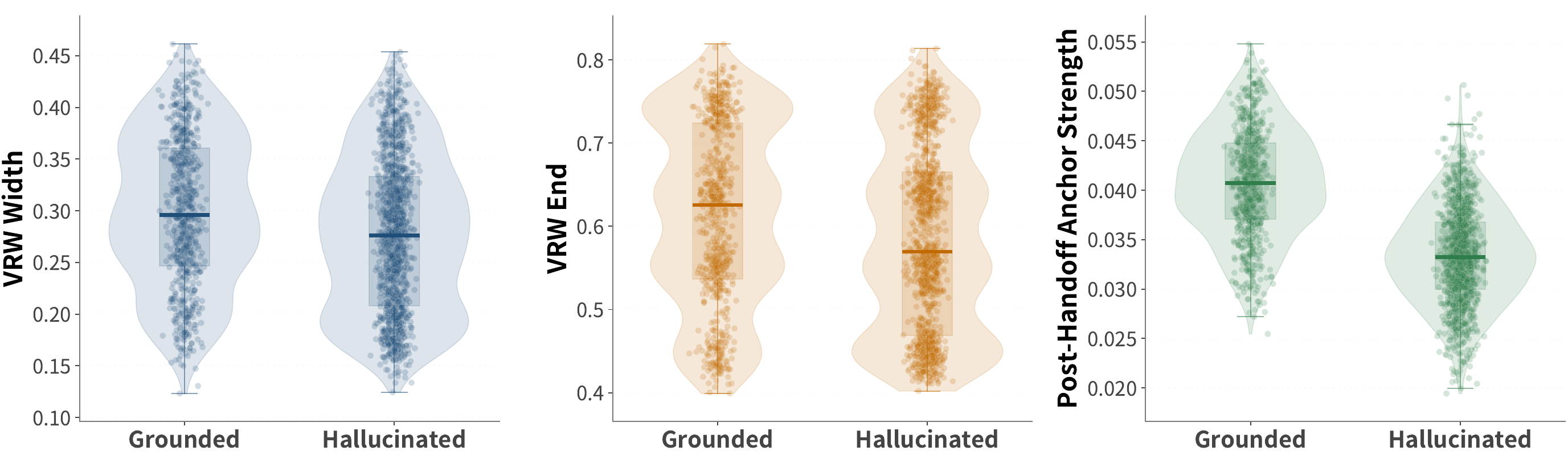}
\caption{Unsupported answers are associated with narrower relay windows, earlier relay termination, and weaker post-handoff visual anchoring than grounded samples on Qwen3-VL-4B.}
\label{fig:hallucination_stats}
\end{figure}

Figure~\ref{fig:hallucination_case} shows a representative split. The grounded branch sustains a broader relay phase and stronger visual anchoring after relay. The same pattern appears at the population level: Figure~\ref{fig:hallucination_stats} shows that unsupported samples concentrate in a regime of narrower relay windows, earlier relay termination, and weaker post-handoff anchoring ($t$-test, all $p<0.01$). These case-level and population-level differences establish a consistent correlation.

To test whether this interval is causally important, we construct grounded--unsupported continuation pairs from repeated decodes of the same input and define \(t_{\mathrm{div}}\) as the first generated token position at which the two answers diverge. For each pair, we treat the grounded branch as the source branch and the unsupported branch as the target branch. At all aligned decoding steps \(t \ge t_{\mathrm{div}}\), we patch the target branch by replacing the visual-token residual activations at layers in a selected range \(\mathcal{L}\) with the corresponding activations from the source branch. We compare the estimated VRW with the full pre-relay and post-relay ranges, as well as with same-width shifted controls. For each sample, let \([s,e]\) denote the estimated VRW and \(w=e-s+1\) its width. We also evaluate width-matched shifted windows of the form \([s+\Delta,e+\Delta]\) with \(\Delta \in \{-2,-1,+1,+2\}\).
Table~\ref{tab:relay_patching_controls} reports the recovery rates, defined as the fraction of unsupported target branches that are reassigned to grounded after patching. Patching the estimated VRW recovers grounded responses in 41.4\% of unsupported cases, compared with 8.7\% for the pre-relay range and 11.2\% for the post-relay range. The same advantage persists after averaging over same-width shifts, indicating that the effect is specific to relay geometry rather than a nearby window of comparable size, validating VRW as a functionally important middle-layer interval for grounded generation.

\begin{table}[t]
\centering
\caption{Relay patching controls. Pre- and post-relay denote the full ranges before and after the estimated VRW, while shifted controls preserve the VRW width. Recovery denotes the fraction of unsupported branches that become supported after patching.} 
\label{tab:relay_patching_controls}
\footnotesize
\setlength{\tabcolsep}{4.5pt}
\renewcommand{\arraystretch}{1.04}
\begin{tabular}{lc}
\toprule
Patched range & Recovery (\%) \\
\midrule
Pre-relay range & 8.7 \\
Post-relay range & 11.2 \\
Mean over same-width shifts & 27.6 $\pm$ 3.3 \\
\textbf{Estimated VRW} & \textbf{41.4} \\
\bottomrule
\end{tabular}
\end{table}

\subsection{Thinking Models Improve Relay Match and Anchoring}
\label{sec:thinking_relay}
The analyses above suggest a capability-level question: why are stronger reasoning models stronger? We examine this by comparing the Instruct and Thinking variants of Qwen3-VL-4B and 8B. For each task family \(t\), let \(\mu_t^{(w)}\) and \(\mu_t^{(e)}\) denote the family mean of \VRWWidth and \VRWEnd under the Instruct condition. For a sample \(i\) in family \(t(i)\), we define task-conditioned relay mismatch as
\begin{equation}
\scriptsize
\Delta^{(i)}_{\mathrm{width}}
=
\left|\textit{VRW-Width}^{(i)}-\mu_{t(i)}^{(w)}\right|,
\qquad
\Delta^{(i)}_{\mathrm{end}}
=
\left|\textit{VRW-End}^{(i)}-\mu_{t(i)}^{(e)}\right|.
\end{equation}
Lower values indicate that a sample's relay geometry is closer to the task-typical regime.

\begin{table}[t]
\centering
\caption{Relay match and anchoring in Instruct and Thinking variants. Entries report Instruct \(\rightarrow\) Thinking.}
\label{tab:thinking_relay}
\scriptsize
\setlength{\tabcolsep}{1.5pt}
\renewcommand{\arraystretch}{1.02}
\begin{tabular}{lccc}
\toprule
Model & Width dev. $\downarrow$ & End dev. $\downarrow$ & Anchor $\uparrow$ \\
\midrule
Qwen3-VL-4B & $0.084 \rightarrow 0.053$ & $0.061 \rightarrow 0.048$ & $0.031 \rightarrow 0.074$ \\
Qwen3-VL-8B & $0.059 \rightarrow 0.030$ & $0.106 \rightarrow 0.074$ & $0.067 \rightarrow 0.081$ \\
\bottomrule
\end{tabular}
\end{table}

Table~\ref{tab:thinking_relay} shows a consistent pattern across both scales. Thinking variants reduce both relay-geometry deviations, indicating better alignment with the task-typical relay regime. They also increase post-handoff anchoring, suggesting stronger retention of the evidence assembled during relay. The gain does not come from uniformly widening the relay window. Instead, Thinking models improve both relay matching and evidence carryover into answer formation. A plausible explanation is that Thinking-style training encourages delayed answer commitment and more explicit use of intermediate evidence, favoring more stable relay allocation and stronger post-handoff visual support.

\begin{figure*}[t]
    \centering
    \includegraphics[width=\textwidth]{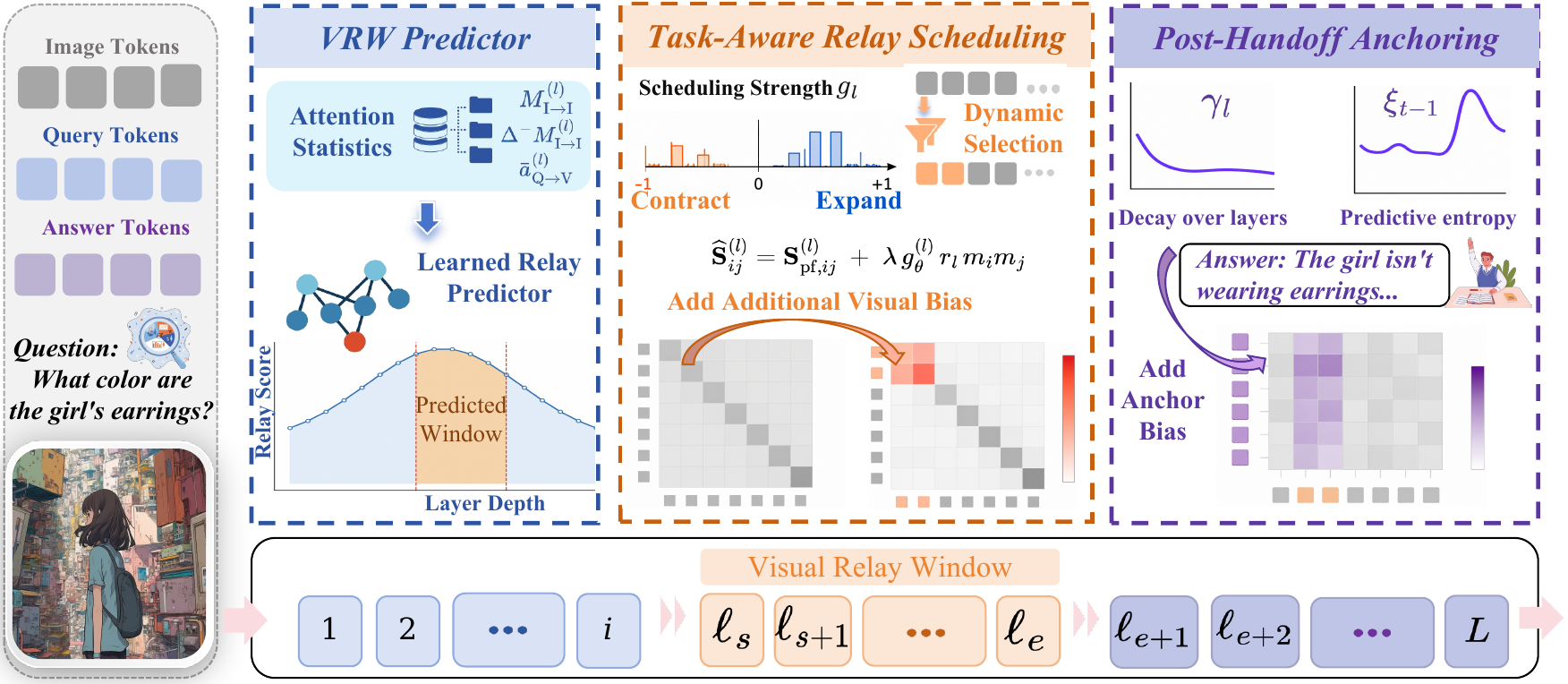}
    \caption{
    Overview of \textbf{TRACE}. TRACE first predicts a task-specific Visual Relay Window from attention statistics. Within the predicted window, it dynamically schedules visual-to-visual attention bias to reinforce evidence relay. After the relay stage ends, it gradually shifts to answer-to-visual anchoring with a depth-decayed bias, preserving grounded visual evidence while supporting stable multimodal reasoning throughout generation.
    }
    \label{fig:trace_overview}
\end{figure*}

\subsection{TRACE: Instantiating Relay-Aware Control}
\label{sec:method}
Building on the relay geometry identified above, Figure~\ref{fig:trace_overview} presents an overview of TRACE, which instantiates this target through three components: a learned relay predictor, a task-aware prefill scheduler with dynamic evidence selection, and a post-handoff anchoring module.

\subsection{A Learned Relay Predictor}
\label{sec:predictor}

A lightweight relay predictor \(D_{\phi}\) assigns each prefill layer a relay score. Causally, the relay score for layer \(l\) is predicted from the statistics observed after layer \(l-1\), namely \(p_l = D_{\phi}\!\left(\mathbf{h}^{(l-1)}\right)\) for \(l \ge 2\), with \(p_1=0\). It operates on compact prefill-time statistics and recovers the relay geometry identified above. The feature vector \(\mathbf{h}^{(l)}\) contains the visual self-attention mass \(M^{(l)}_{\mathrm{I}\rightarrow\mathrm{I}}\), its layerwise change \(\Delta^- M^{(l)}_{\mathrm{I}\rightarrow\mathrm{I}}\), the mean question-to-visual attention \(\bar{a}^{(l)}_{\mathrm{Q}\rightarrow\mathrm{V}}\), and the relative layer depth \(\rho_l\):
\begin{equation}
\mathbf{h}^{(l)}
=
\Big[
M^{(l)}_{\mathrm{I}\rightarrow\mathrm{I}},
\Delta^- M^{(l)}_{\mathrm{I}\rightarrow\mathrm{I}},
\bar{a}^{(l)}_{\mathrm{Q}\rightarrow\mathrm{V}},
\rho_l
\Big],
\end{equation}
where
\(
\bar{a}^{(l)}_{\mathrm{Q}\rightarrow\mathrm{V}}
=
\frac{1}{|\mathcal{Q}|}\sum_{i \in \mathcal{Q}}\sum_{j \in \mathcal{V}}\mathbf{A}^{(l)}_{ij}
\)
and
\(
\Delta^- M^{(l)}_{\mathrm{I}\rightarrow\mathrm{I}} = M^{(l)}_{\mathrm{I}\rightarrow\mathrm{I}} - M^{(l-1)}_{\mathrm{I}\rightarrow\mathrm{I}}.
\)
We implement \(D_{\phi}\) as a two-layer MLP with sigmoid output.

The predictor is trained from pseudo labels induced by the offline VRW definition in Section~\ref{sec:operational_vrw}. For each image-question pair, the rule-based estimator yields a relay interval \([s,e]\), which is converted into layerwise labels
\begin{equation}
z_l =
\begin{cases}
1, & l \in [s,e], \\
0, & \text{otherwise}.
\end{cases}
\end{equation}
We then optimize
\begin{equation}
\mathcal{L}_{\mathrm{det}}
=
\frac{1}{L}\sum_{l=1}^{L}
\mathrm{BCE}(p_l, z_l).
\end{equation}
This objective trains the predictor to recover relay structure from prefill-time signals alone.

\subsection{Task-Aware Relay Scheduling}
\label{sec:scheduler}

Given the relay profile, the scheduler determines where to intervene and which visual content to preserve.

\paragraph{Layerwise scheduling}
Let \(\mathbf{S}^{(l)}_{\mathrm{pf}}\) denote the pre-softmax attention logits at prefill layer \(l\). From the predictor output we derive a soft relay gate \(r_l = \sigma\!\big(\alpha (p_l - \beta)\big)\), where \(\alpha\) controls sharpness and \(\beta\) specifies the operating point. A layerwise scheduling strength \(g_{\theta}^{(l)} \in [-1,1]\) is then computed by conditionally modulating the relay context \(\mathbf{c}^{(l)}\) with the mean-pooled previous-layer  representation of question tokens \(\mathbf{q}\). Here \(\mathbf{c}^{(l)}\) contains \(\rho_l\), the predictor confidence \(p_l\), and the local relay statistics above. The question branch produces feature-wise scaling terms,
\begin{equation}
[\boldsymbol{\gamma}_q,\boldsymbol{\beta}_q] = W_q \mathbf{q},
\qquad
\widetilde{\mathbf{c}}^{(l)} = \boldsymbol{\gamma}_q \odot \mathbf{c}^{(l)} + \boldsymbol{\beta}_q,
\end{equation}
and the scheduler predicts
\begin{equation}
g_{\theta}^{(l)} = \tanh\!\big(\mathrm{MLP}(\widetilde{\mathbf{c}}^{(l)})\big).
\end{equation}
Positive values of \(g_{\theta}^{(l)}\) broaden or strengthen visual consolidation, while negative values contract over-extended relay phases and move handoff earlier. 

\paragraph{Dynamic evidence selection}
Within predicted relay layers, the scheduler maintains a task-relevant visual support set, concentrating the intervention on the subset of visual content currently recruited by the question. At each relay-active layer, question-to-visual attention defines a relevance score
\begin{equation}
a^{(l)}_j
=
\frac{1}{|\mathcal{Q}|}
\sum_{i \in \mathcal{Q}}
\mathbf{A}^{(l)}_{ij},
\qquad j \in \mathcal{V},
\end{equation}
which updates a running score via \(u^{(l)}_j = (1-\mu)u^{(l-1)}_j + \mu a^{(l-1)}_j\) for \(l \ge 2\), with \(u^{(1)}_j=0\). We then keep the top-\(p\) fraction of visual tokens:
\begin{equation}
S_{\mathrm{vis}}^{(l)} = \mathrm{TopP}(\mathbf{u}^{(l)}, p),
\end{equation}
During prefill, we maintain an online support mask \(m_j^{(l)} = \mathbb{I}[j \in S_{\mathrm{vis}}^{(l)}]\) updated from accumulated question-to-visual relevance and apply it immediately at layer \(l\). After prefill, we freeze the final support set \(S_{\mathrm{vis}}\) for decode-time anchoring, with corresponding mask \(m_j = \mathbb{I}[j \in S_{\mathrm{vis}}]\).

During prefill, the scheduler directly reshapes visual-to-visual interactions in relay layers:
\begin{equation}
\widehat{\mathbf{S}}^{(l)}_{ij}
=
\mathbf{S}^{(l)}_{\mathrm{pf},ij}
\;+\;
\lambda\, g_{\theta}^{(l)}\, r_l\, m_i^{(l)} m_j^{(l)},
\qquad i,j \in \mathcal{V},
\end{equation}
where \(\lambda\) is a global prefill scheduling scale. All other entries are left unchanged. The resulting bias broadens, contracts, or stabilizes the relay phase according to task demand before answer generation begins.

\subsection{Post-Handoff Anchoring for Evidence Preservation}
\label{sec:anchoring}
Relay shaping alone is not sufficient: once the relay phase ends, the model must still retain stable access to the visual support it has already assembled. We therefore introduce a post-handoff anchoring mechanism that carries the selected support into answer generation. Using the relay gate \(r_l\), we identify relay exit from the prefill profile with a thresholded stopping rule, and take the deepest layer with sustained relay confidence as the handoff point:
\begin{equation}
\hat{e} = \max \{\, l \mid r_l \ge \tau_r \,\}.
\end{equation}
The frozen support set \(S_{\mathrm{vis}}\) is then carried into the post-handoff stage. We define
\begin{equation}
\mathrm{Anchor}^{(l)}
=
\frac{1}{|\mathcal{Y}|}
\sum_{i \in \mathcal{Y}}
\frac{\sum_{j \in S_{\mathrm{vis}}}\mathbf{A}^{(l)}_{ij}}
{\sum_{k \in \mathcal{V}}\mathbf{A}^{(l)}_{ik} + \epsilon},
\end{equation}
which measures how much of the answer-side visual attention remains concentrated on task-relevant content. During training, \(\mathcal{Y}\) denotes the teacher-forced target answer positions; at inference, the same anchoring bias is applied one decoding step at a time.

Anchoring is implemented as a weak additive bias on decode-time answer-to-visual logits. The bias is strongest immediately after handoff and then gradually relaxes in deeper layers. We further modulate it by the log-normalized predictive entropy from the previous decoding step, denoted \(\xi_{t-1} \in [0,1]\), so that uncertain generations retain stronger access to the visual support. 

Let \(\mathbf{S}^{(l)}_{\mathrm{dec}}\) denote the native decode-time logits at layer \(l\), with \(j\) indexing cached visual tokens. During training, anchoring uses the soft post-handoff weight \(1-r_l\); at inference, the same bias is applied stepwise with the hard mask \(\mathbb{I}[l>\hat e]\):
\begin{equation}
\small
\widetilde{\mathbf{S}}^{(l)}_{\mathrm{dec},ij}
=
\mathbf{S}^{(l)}_{\mathrm{dec},ij}
\;+\;
\omega_l\,\gamma_l\,(1+\xi_{t-1})\, m_j,
\qquad
j \in \mathcal{V},
\end{equation}
where \(\omega_l = 1-r_l\) during training and \(\omega_l = \mathbb{I}[l>\hat e]\) at inference, and \(\gamma_l = \gamma_0 \exp(-\kappa (l/L))\) with fixed \(\gamma_0,\kappa > 0\).

\subsection{Initialization and Joint Optimization}
\label{sec:training}

We first warm-start the relay predictor with \(\mathcal{L}_{\mathrm{det}}\), and then jointly optimize the predictor, scheduler, and anchoring module on supervised image-question-answer triples, while keeping the base VLM parameters frozen. During training, discrete support selection is replaced by a differentiable soft mask and post-handoff behavior is weighted by the soft complement \(1-r_l\); at inference, these relaxations revert to the hard top-\(p\) support set and the handoff point \(\hat e\). The overall objective is
\begin{equation}
\mathcal{L}_{\mathrm{int}}
=
\mathcal{L}_{\mathrm{task}}
\;+\;
\lambda_{\mathrm{anchor}} \mathcal{L}_{\mathrm{anchor}},
\end{equation}
where \(\mathcal{L}_{\mathrm{task}}\) is the standard next-token objective and \(\mathcal{L}_{\mathrm{anchor}}\) encourages post-handoff evidence retention during answer formation:
\begin{equation}
\mathcal{L}_{\mathrm{anchor}}
=
\frac{1}{L}
\sum_{l=1}^{L}
\left(1-r_l\right)\max\!\big(0, \eta - \mathrm{Anchor}^{(l)}\big).
\end{equation}

\section{Experiments}
\label{sec:experiments}

\begin{table*}[t]
\centering
\caption{Main benchmark results across four model backbones. Values in parentheses indicate absolute change relative to the corresponding vanilla model. The best result in each model--benchmark row is shown in bold, and the second-best result is underlined. The \textit{Average} row reports the unweighted mean over the seven benchmarks for each backbone.}
\label{tab:main_results}
\scriptsize
\setlength{\tabcolsep}{3.8pt}
\renewcommand{\arraystretch}{1.10}
\begin{tabular}{>{\raggedright\arraybackslash}p{1.55cm}>{\raggedright\arraybackslash}p{1.95cm}>{\columncolor{tablevanilla}}c c c c c c c c >{\columncolor{tableours}}c}
\toprule
\rowcolor{tablegray}
Model & Benchmark & \cellcolor{tablevanillahead}\shortstack{Vanilla} & \shortstack{Best-of-\(N\)\\{\footnotesize arXiv'22}} & \shortstack{CoT\\{\footnotesize NeurIPS'22}} & \shortstack{VCD\\{\footnotesize CVPR'24}} & \shortstack{PAI\\{\footnotesize ECCV'24}} & \shortstack{MARINE\\{\footnotesize arXiv'24}} & \shortstack{PTI\\{\footnotesize CVPR'26}} & \shortstack{DMAS\\{\footnotesize ICLR'26}} & \cellcolor{tableourshead}\shortstack{TRACE\\{\footnotesize Ours}} \\
\midrule
\multirow{7}{*}{\shortstack{\textbf{Qwen3-VL-4B}}} & DocVQA& \score{95.3}{} & \score{94.8}{-0.5} & \bestscore{95.9}{+0.6} & \score{94.9}{-0.4} & \score{94.8}{-0.5} & \score{93.5}{-1.8} & \score{95.4}{+0.1} & \score{94.7}{-0.6} & \secondscore{95.6}{+0.3} \\
& ChartQA & \score{84.6}{} & \score{84.9}{+0.3} & \secondscore{85.1}{+0.5} & \score{84.2}{-0.4} & \score{84.6}{+0.0} & \score{84.8}{+0.2} & \score{84.8}{+0.2} & \score{84.9}{+0.3} & \bestscore{87.3}{+2.7} \\
& MathVista& \score{73.7}{} & \bestscore{77.3}{+3.6} & \score{75.8}{+2.1} & \score{73.3}{-0.4} & \score{74.1}{+0.4} & \score{74.6}{+0.9} & \score{74.4}{+0.7} & \score{76.0}{+2.3} & \secondscore{76.8}{+3.1} \\
& RealWorldQA & \score{70.9}{} & \score{73.3}{+2.4} & \score{72.2}{+1.3} & \score{71.1}{+0.2} & \score{71.7}{+0.8} & \score{71.6}{+0.7} & \secondscore{73.4}{+2.5} & \score{72.0}{+1.1} & \bestscore{73.7}{+2.8} \\
& HallusionBench & \score{57.6}{} & \score{57.0}{-0.6} & \score{56.8}{-0.8} & \score{60.9}{+3.3} & \score{62.5}{+4.9} & \score{62.1}{+4.5} & \score{62.7}{+5.1} & \secondscore{63.0}{+5.4} & \bestscore{63.4}{+5.8} \\
& VlmsAreBlind & \score{71.9}{} & \score{72.4}{+0.5} & \score{71.2}{-0.7} & \score{75.3}{+3.4} & \score{76.3}{+4.4} & \score{76.1}{+4.2} & \secondscore{76.4}{+4.5} & \score{76.1}{+4.2} & \bestscore{76.8}{+4.9} \\
& CountBench& \score{84.9}{} & \score{85.4}{+0.5} & \score{84.1}{-0.8} & \score{85.9}{+1.0} & \score{87.4}{+2.5} & \bestscore{88.2}{+3.3} & \secondscore{87.6}{+2.7} & \score{86.9}{+2.0} & \secondscore{87.6}{+2.7} \\
\cmidrule[0.3pt](lr){2-11}
& \textbf{Average} & \score{76.99}{} & \score{77.87}{+0.88} & \score{77.30}{+0.31} & \score{77.94}{+0.95} & \score{78.77}{+1.78} & \score{78.70}{+1.71} & \secondscore{79.24}{+2.25} & \score{79.09}{+2.10} & \bestscore{80.17}{+3.18} \\
\midrule
\multirow{7}{*}{\shortstack{\textbf{Qwen3-VL-8B}}} & DocVQA & \score{96.1}{} & \score{96.0}{-0.1} & \score{95.4}{-0.7} & \secondscore{96.5}{+0.4} & \score{95.6}{-0.5} & \score{95.7}{-0.4} & \score{96.1}{+0.0} & \score{95.6}{-0.5} & \bestscore{96.7}{+0.6} \\
& ChartQA & \score{89.6}{} & \score{89.4}{-0.2} & \score{89.0}{-0.6} & \score{88.8}{-0.8} & \score{89.1}{-0.5} & \score{89.2}{-0.4} & \score{89.3}{-0.3} & \bestscore{90.5}{+0.9} & \secondscore{90.1}{+0.5} \\
& MathVista & \score{77.2}{} & \bestscore{81.6}{+4.4} & \score{79.1}{+1.9} & \score{77.2}{+0.0} & \score{77.9}{+0.7} & \score{77.6}{+0.4} & \score{78.6}{+1.4} & \score{79.1}{+1.9} & \secondscore{80.0}{+2.8} \\
& RealWorldQA& \score{71.5}{} & \secondscore{73.4}{+1.9} & \score{73.0}{+1.5} & \score{71.5}{+0.0} & \score{72.5}{+1.0} & \score{72.2}{+0.7} & \score{72.4}{+0.9} & \score{72.8}{+1.3} & \bestscore{74.2}{+2.7} \\
& HallusionBench & \score{61.1}{} & \score{61.3}{+0.2} & \score{60.7}{-0.4} & \score{64.4}{+3.3} & \score{65.0}{+3.9} & \score{65.4}{+4.3} & \score{65.2}{+4.1} & \secondscore{65.8}{+4.7} & \bestscore{66.1}{+5.0} \\
& VlmsAreBlind & \score{74.0}{} & \score{74.4}{+0.4} & \score{74.3}{+0.3} & \score{76.3}{+2.3} & \score{78.1}{+4.1} & \score{78.0}{+4.0} & \bestscore{78.5}{+4.5} & \secondscore{78.4}{+4.4} & \bestscore{78.5}{+4.5} \\
& CountBench& \score{80.5}{} & \score{82.1}{+1.6} & \score{80.1}{-0.4} & \secondscore{84.7}{+4.2} & \score{84.0}{+3.5} & \bestscore{84.9}{+4.4} & \score{84.3}{+3.8} & \secondscore{84.7}{+4.2} & \score{84.4}{+3.9} \\
\cmidrule[0.3pt](lr){2-11}
& \textbf{Average} & \score{78.57}{} & \score{79.74}{+1.17} & \score{78.80}{+0.23} & \score{79.91}{+1.34} & \score{80.31}{+1.74} & \score{80.43}{+1.86} & \secondscore{80.63}{+2.06} & \score{80.99}{+2.42} & \bestscore{81.43}{+2.86} \\
\midrule
\multirow{7}{*}{\shortstack{\textbf{InternVL3.5-4B}}} & DocVQA& \score{92.4}{} & \secondscore{92.9}{+0.5} & \bestscore{93.2}{+0.8} & \score{91.8}{-0.6} & \score{92.0}{-0.4} & \score{92.1}{-0.3} & \score{92.2}{-0.2} & \score{92.6}{+0.2} & \score{92.7}{+0.3} \\
& ChartQA& \score{86.2}{} & \score{87.1}{+0.9} & \bestscore{87.8}{+1.6} & \score{86.0}{-0.2} & \score{87.2}{+1.0} & \score{86.3}{+0.1} & \score{86.3}{+0.1} & \score{86.5}{+0.3} & \secondscore{87.5}{+1.3} \\
& MathVista& \score{71.4}{} & \secondscore{74.6}{+3.2} & \score{72.6}{+1.2} & \score{71.1}{-0.3} & \score{72.1}{+0.7} & \score{71.9}{+0.5} & \score{72.3}{+0.9} & \score{73.6}{+2.2} & \bestscore{74.9}{+3.5} \\
& RealWorldQA & \score{66.3}{} & \score{67.9}{+1.6} & \score{67.6}{+1.3} & \score{66.7}{+0.4} & \score{67.3}{+1.0} & \score{67.3}{+1.0} & \score{67.9}{+1.6} & \secondscore{68.2}{+1.9} & \bestscore{70.3}{+4.0} \\
& HallusionBench~\cite{guan2024hallusionbench} & \score{44.8}{} & \score{43.1}{-1.7} & \score{44.3}{-0.5} & \score{47.8}{+3.0} & \score{48.4}{+3.6} & \secondscore{49.1}{+4.3} & \score{48.8}{+4.0} & \score{49.0}{+4.2} & \bestscore{49.3}{+4.5} \\
& VlmsAreBlind& \score{64.8}{} & \score{65.2}{+0.4} & \score{64.9}{+0.1} & \score{68.3}{+3.5} & \score{68.7}{+3.9} & \bestscore{69.2}{+4.4} & \secondscore{69.0}{+4.2} & \score{68.9}{+4.1} & \score{68.3}{+3.5} \\
& CountBench& \score{85.6}{} & \score{86.4}{+0.8} & \score{85.9}{+0.3} & \score{87.0}{+1.4} & \score{88.3}{+2.7} & \bestscore{88.8}{+3.2} & \score{87.4}{+1.8} & \score{88.1}{+2.5} & \secondscore{88.4}{+2.8} \\
\cmidrule[0.3pt](lr){2-11}
& \textbf{Average} & \score{73.07}{} & \score{73.89}{+0.81} & \score{73.76}{+0.69} & \score{74.10}{+1.03} & \score{74.86}{+1.79} & \score{74.96}{+1.89} & \score{74.84}{+1.77} & \secondscore{75.27}{+2.20} & \bestscore{75.91}{+2.84} \\
\midrule
\multirow{7}{*}{\shortstack{\textbf{InternVL3.5-8B}}} & DocVQA& \score{92.3}{} & \bestscore{93.0}{+0.7} & \score{92.0}{-0.3} & \score{92.3}{+0.0} & \score{92.0}{-0.3} & \score{92.1}{-0.2} & \score{91.9}{-0.4} & \score{92.1}{-0.2} & \secondscore{92.7}{+0.4} \\
& ChartQA& \score{86.7}{} & \score{87.1}{+0.4} & \secondscore{87.4}{+0.7} & \score{86.0}{-0.7} & \score{86.8}{+0.1} & \score{87.0}{+0.3} & \score{86.9}{+0.2} & \score{87.0}{+0.3} & \bestscore{88.0}{+1.3} \\
& MathVista & \score{74.2}{} & \bestscore{77.9}{+3.7} & \score{76.4}{+2.2} & \score{73.8}{-0.4} & \score{75.0}{+0.8} & \score{74.2}{+0.0} & \score{75.0}{+0.8} & \score{76.2}{+2.0} & \secondscore{77.0}{+2.8} \\
& RealWorldQA& \score{67.5}{} & \score{69.2}{+1.7} & \secondscore{69.4}{+1.9} & \score{67.8}{+0.3} & \score{68.3}{+0.8} & \score{68.3}{+0.8} & \score{68.8}{+1.3} & \score{68.6}{+1.1} & \bestscore{69.8}{+2.3} \\
& HallusionBench & \score{54.5}{} & \score{54.8}{+0.3} & \score{54.0}{-0.5} & \score{58.1}{+3.6} & \score{59.6}{+5.1} & \score{59.3}{+4.8} & \score{59.2}{+4.7} & \secondscore{59.9}{+5.4} & \bestscore{61.1}{+6.6} \\
& VlmsAreBlind & \score{66.1}{} & \score{66.0}{-0.1} & \score{65.6}{-0.5} & \score{68.5}{+2.4} & \score{69.1}{+3.0} & \score{70.3}{+4.2} & \secondscore{70.8}{+4.7} & \score{70.1}{+4.0} & \bestscore{70.9}{+4.8} \\
& CountBench& \score{86.2}{} & \score{87.1}{+0.9} & \score{87.0}{+0.8} & \score{88.5}{+2.3} & \score{88.3}{+2.1} & \bestscore{89.3}{+3.1} & \score{89.0}{+2.8} & \score{89.0}{+2.8} & \secondscore{89.2}{+3.0} \\
\cmidrule[0.3pt](lr){2-11}
& \textbf{Average} & \score{75.36}{} & \score{76.44}{+1.08} & \score{75.97}{+0.61} & \score{76.43}{+1.07} & \score{77.01}{+1.65} & \score{77.21}{+1.85} & \score{77.37}{+2.01} & \secondscore{77.56}{+2.20} & \bestscore{78.39}{+3.03} \\
\bottomrule
\end{tabular}
\end{table*}

\subsection{Models, Datasets, and Evaluation Protocols}
\label{sec:exp_setup}

  We evaluate four representative vision-language models from two mainstream families: Qwen3-VL-4B, Qwen3-VL-
  8B\cite{yang2025qwen3}, InternVL3.5-4B, and InternVL3.5-8B\cite{wang2025internvl3}. Each backbone is trained separately
  with its own TRACE modules. Our evaluation covers seven benchmarks: HallusionBench\cite{guan2024hallusionbench},VlmsAreBlind\cite{rahmanzadehgervi2024vision}, and CountBench\cite{amini2024countgd} for grounding-sensitive settings;
  DocVQA\cite{mathew2021docvqa}, ChartQA\cite{masry2022chartqa}, and RealWorldQA\cite{grok15v2024} for fine-grained evidence settings;
  and MathVista\cite{lu2024mathvista} for reasoning-heavy settings. We compare against eight baselines: vanilla inference,
  self-consistency / best-of-\(N\) (\(N\)=10, temperature=0.7)\cite{wang2022self}, chain-of-thought prompting
  (CoT)\cite{wei2022chain}, Visual Contrastive Decoding (VCD)\cite{leng2024mitigating}, Paying More Attention to Image
  (PAI)\cite{liu2024paying}, MARINE\cite{zhao2024mitigating}, Prefill-Time Intervention (PTI)\cite{zhang2026prefill}, and
  Dynamic Multimodal Activation Steering (DMAS)\cite{yin2026dynamic}.

The relay predictor is trained on 2{,}000 image-question pairs, constructed by uniformly sampling 500 examples each from Visual Genome\cite{krishna2017visual}, VQA\cite{antol2015vqa}, AI2D\cite{hiippala2021ai2d}, and ScienceQA\cite{lu2022learn} to respectively cover the four relay-demand categories; pseudo labels are induced automatically by the VRW estimator in Section~\ref{sec:operational_vrw}. The full intervention module is then optimized on 20{,}000 samples from  FineVision\cite{wiedmann2026finevisionopendataneed}. We exclude training samples from datasets that overlap with downstream benchmarks to avoid data contamination.

\begin{table}[t]
\centering
\caption{Trainable TRACE modules.}
\label{tab:trace_modules}
\scriptsize
\setlength{\tabcolsep}{0pt}
\renewcommand{\arraystretch}{1.05}
\begin{tabular}{p{1.1cm}p{3.6cm}p{1cm}cc}
\toprule
\rowcolor{tablegray}
Module & Structure & Hidden size & Params (4B) & Params (8B) \\
\midrule
Predictor & 2-layer MLP on 4-d relay statistics & 128 & 17,281 & 17,281 \\
Scheduler & Question projection, affine modulation, 2-layer MLP control head & 128 & 181,963 & 280,267 \\
Anchoring & Structured decode-time bias  & -- & 0 & 0 \\
\midrule
\textbf{Total} & \multicolumn{2}{l}{\textbf{TRACE controller}} & \textbf{199,244} & \textbf{297,548} \\
\bottomrule
\end{tabular}
\end{table}
Table~\ref{tab:trace_modules} summarizes the trainable TRACE components and their parameter budgets. The predictor size is constant across backbones. The scheduler differs only in the dimension of the pooled question representation, which is 2560 for the 4B backbones and 4096 for the 8B backbones. We report standard accuracy metrics and implement all baselines using their official codebases. We use greedy decoding with \( \mathrm{temperature}=0 \), \( \mathrm{do\_sample}=\mathrm{False} \), and \( \mathrm{max\_new\_tokens}=4096 \). Visual inputs are processed with each model's official processor under dynamic-resolution settings. We retain the top \(p=0.3\) fraction of visual tokens in the support set, use a relay-exit threshold of \(\tau_r=0.7\), and set the post-handoff evidence-retention loss to \(\lambda_{\mathrm{anchor}}=0.05\). Other control hyperparameters are fixed: \(\alpha=13\), \(\beta=0.5\), \(\mu=0.15\), \(\eta=0.7\), \(\lambda=0.6\), \(\gamma_0=0.4\), and \(\kappa=2.0\). We train for 3 epochs in Stage~1 and 2 epochs in Stage~2 on four NVIDIA RTX 5090 GPUs. The full training process takes approximately 8 hours, and the base VLM parameters remain frozen throughout. We use separate optimization settings for the two stages. Stage~1 warm-starts the relay predictor with learning rate $1\times 10^{-4}$ and batch size 32. Stage~2 performs joint optimization with learning rate $5\times 10^{-5}$ and batch size 64. Both stages use AdamW and a linear warmup over the first 100 steps.
\subsection{Main Benchmark Results}
\label{sec:main_results}
Table~\ref{tab:main_results} summarizes the main benchmark results across four model backbones. TRACE improves performance consistently. The clearest improvements appear on grounding-sensitive settings: averaged over HallusionBench, VlmsAreBlind, and CountBench across all four backbones, TRACE improves over vanilla by 4.33 points, with a largest single gain of 6.6 points on HallusionBench for InternVL3.5-8B. On the fine-grained settings DocVQA, ChartQA, and RealWorldQA, TRACE yields a mean gain of 1.60 points. On the reasoning-heavy setting MathVista, it improves over vanilla on all four backbones, with an average gain of 3.05 points.

Across all 28 model--benchmark pairs, TRACE improves over vanilla by 2.98 points on average. Overall, Best-of-\(N\) and CoT are strongest when explicit search is especially helpful, while VCD, PAI, MARINE, PTI, and DMAS are more competitive on visually sensitive benchmarks. Against these strong baselines, TRACE maintains a strong overall trade-off across grounded generation, reasoning, and fine-grained evidence extraction.

\subsection{Mechanism-Level Analysis of Relay Reshaping}
\label{sec:mechanism_analysis}

\begin{figure}
    \centering
    \includegraphics[width=\linewidth]{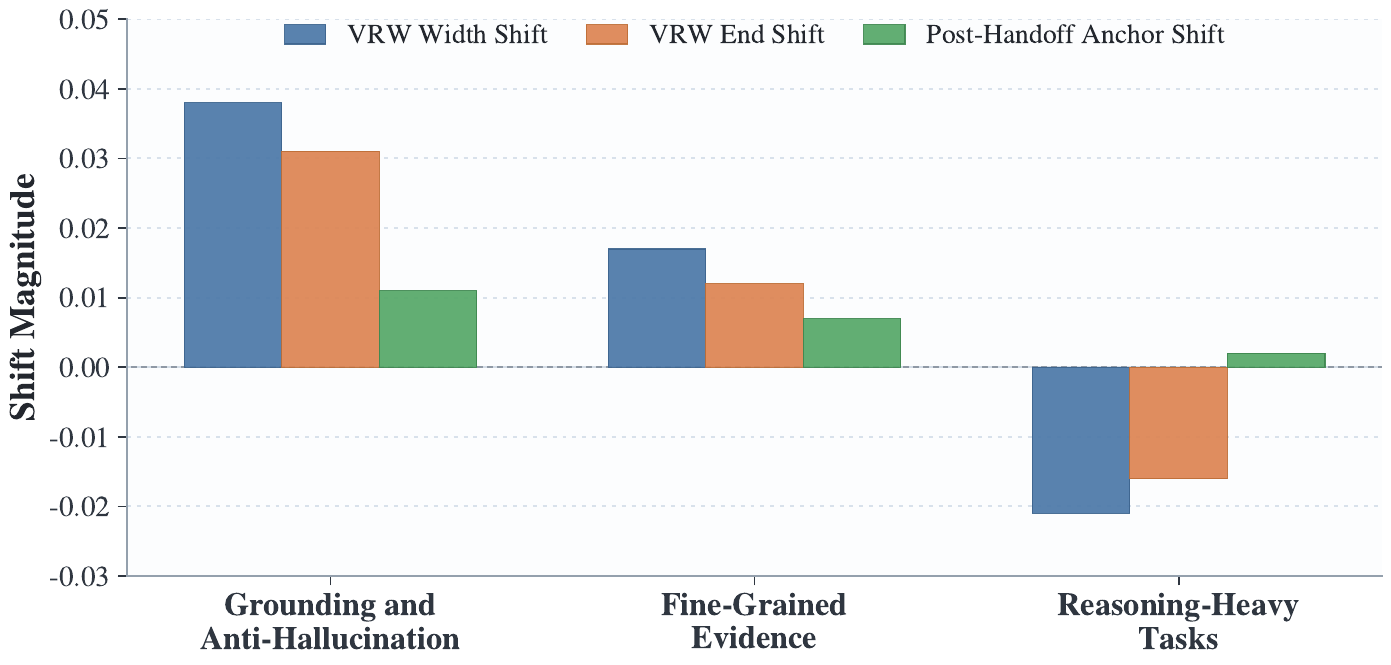}
    \caption{Task-wise relay reshaping on Qwen3-VL-4B. Positive values indicate a broader or later relay phase and stronger post-
    handoff visual retention.}
    \label{fig:taskwise_relay_shift}
\end{figure}

Beyond benchmark gains, we ask whether the intervention reshapes computation in the intended way. Figure~\ref{fig:taskwise_relay_shift} shows a clear task dependence. On grounding-sensitive benchmarks including HallusionBench and VlmsAreBlind, the intervention broadens the relay phase and delays handoff. On DocVQA, ChartQA and RealWorldQA the shift is milder but still positive, reflecting the need to preserve fine-grained local evidence through final answer generation. On reasoning-heavy benchmarks such as MathVista, the pattern reverses: the relay window often becomes shorter and ends earlier, leaving more depth for language-side reasoning once the relevant visual evidence has already been organized. In other words, TRACE improves performance by reshaping the same relay redistribution pattern identified in analysis, rather than by uniformly amplifying visual influence.

\subsection{Comparison with Lightweight Learned Adaptation}
\label{sec:learned_adaptation}

We further test whether TRACE helps simply by adding a small amount of trainable adaptation. We compare TRACE with two lightweight learned baselines under the same training pipeline. The first baseline is a {Generic Controller}. It retains the same intervention interface as TRACE, but removes the relay-specific structure.  Concretely, it predicts one per-layer prefill bias and one per-layer decode bias from the pooled  question representation and relative layer depth through a shared controller with a 64-dimensional question projection, a 64-dimensional layer embedding, a 128-dimensional hidden state, and two scalar output heads. This gives 199,746 trainable parameters on 4B backbones and 298,050 on 8B backbones, closely matching TRACE. The second is an \emph{Attention-LoRA} variant. It inserts LoRA adapters into the \(q\)- and \(v\)-projection matrices of all self-attention layers while keeping the backbone otherwise frozen. We use rank \(r=4\) for the 4B backbones and \(r=8\) for the 8B backbones.

\begin{table}[t]
\centering
\caption{Comparison with lightweight learned adaptation on representative benchmarks.}
\label{tab:learned_adaptation}
\scriptsize
\setlength{\tabcolsep}{3.3pt}
\renewcommand{\arraystretch}{1.04}
\begin{tabular}{lccccc}
\toprule
\rowcolor{tablegray}
Method & HallusionBench & VlmsAreBlind & MathVista & DocVQA & Avg. \\
\midrule
Vanilla & 57.6 & 71.9 & 73.7 & 95.3 & 74.6 \\
Generic Controller & \underline{61.9} & 71.0 & \underline{75.6} & 95.1 & 75.9 \\
Attention-LoRA & {61.6} & \underline{75.8} & 72.1 & \underline{95.4} & \underline{76.2} \\
\rowcolor{tableours}
TRACE & \textbf{63.4} & \textbf{76.8} & \textbf{76.8} & \textbf{95.6} & \textbf{78.2} \\
\bottomrule
\end{tabular}
\end{table}

Table~\ref{tab:learned_adaptation} shows that lightweight learned adaptation helps, but does not match TRACE. The Generic Controller improves over vanilla by only 1.3 points on average while  Attention-LoRA remains 2.0 points below TRACE. This indicates that the remaining advantage of TRACE lies not in adaptation capacity alone, but in how that capacity is  organized around model's internal structure.

\subsection{Component Ablations and Control Sensitivity}
\label{sec:ablation}
We ablate the three major components as well as two auxiliary design choices, and vary the key control hyperparameters \(p\), \(\tau_r\), and \(\lambda_{\mathrm{anchor}}\). For `w/o Predictor', we replace the learned predictor with the rule-based VRW estimator via a preliminary forward pass; for `w/o Scheduler', we remove the prefill-time \(V\!\to\!V\) relay bias; for `w/o Anchoring', we remove both the decode-time anchoring bias and the anchoring loss; for  `Positive-only scheduler', we clamp negative scheduler outputs to zero; for `No entropy modulation', we remove the \((1+\xi_{t-1})\) factor from decode-time anchoring.

\begin{table}[t]
\centering
\caption{Component ablations on Qwen3-VL-4B. The blue row denotes the full model. The best result in each column is shown in bold, and the second-best result is underlined. }
\label{tab:component_ablation}
\scriptsize
\setlength{\tabcolsep}{2.8pt}
\renewcommand{\arraystretch}{1.03}
\begin{tabular}{lccccc}
\toprule
\rowcolor{tablegray}
Method & HallusionBench & VlmsAreBlind & MathVista & DocVQA & Avg. \\
\midrule
Vanilla & 57.6 & 71.9 & 73.7 & 95.3 & 74.6 \\
w/o Predictor & 61.8 & 75.1 & 75.4 & \textbf{95.8} & 77.0 \\
w/o Scheduler & 59.7 & 75.8 & 75.2 & 95.2 & 76.5 \\
w/o Anchoring & 61.4 & 74.7 & \textbf{77.6} & 95.1 & 77.2 \\
{Positive-only scheduler} & \underline{63.0} & \underline{76.2} & 74.6 & 95.3 & {77.3} \\
{No entropy modulation} & 62.8 & 76.0 & 76.4 & 95.4 & \underline{77.7} \\
\rowcolor{tableours}
TRACE (Full) & \textbf{63.4} & \textbf{76.8} & \underline{76.8} & \underline{95.6} & \textbf{78.2} \\
\bottomrule
\end{tabular}
\end{table}

Table~\ref{tab:component_ablation} shows that the three original components play distinct roles. Removing the predictor causes a consistent drop from the full model but still preserves most of the gain over vanilla, because the rule-based VRW estimator provides a usable coarse relay location while the learned predictor mainly improves input-specific stability. Removing the scheduler hurts HallusionBench most directly, as it is the component that reshapes relay before handoff and determines whether enough visual evidence is assembled for grounding-sensitive cases. Removing anchoring leaves MathVista highest among the ablations, but weakens the grounding benchmarks relative to the full model, because anchoring mainly preserves visual support after handoff instead of determining how long relay should last. `Positive-only scheduler' shows that contraction is necessary when later language-side reasoning should take over, while `No entropy modulation' indicates that the dynamic decode-time bias is helpful. Averaged over the four reported benchmarks, the full model remains best.

\begin{table}[t]
\centering
\caption{Sensitivity to key control hyperparameters on Qwen3-VL-4B. The blue row in each block marks the default setting. Within each block, the best value in each column is shown in bold and the second-best value is underlined. }
\label{tab:hyperparam_sensitivity}
\scriptsize
\setlength{\tabcolsep}{1.5pt}
\renewcommand{\arraystretch}{1.04}
\begin{tabular}{lcccccc}
\toprule
\rowcolor{tablegray}
Setting & HallusionBench & VlmsAreBlind & MathVista & DocVQA & Avg. & \AnchorStrength \\
\midrule
\rowcolor{tablegray}
\multicolumn{7}{l}{\textit{Support ratio} $p$} \\
0.15 & 61.4 & 74.9 & 75.0 & 94.2 & 76.4 & 0.066 \\
0.20 & 62.6 & 75.8 & 76.0 & \underline{95.1} & 77.4 & 0.071 \\
\rowcolor{tableours}
0.30 & \underline{63.4} & \textbf{76.8} & \textbf{76.8} & \textbf{95.6} & \textbf{78.2} & 0.074 \\
0.40 & \textbf{63.9} & \underline{76.3} & 76.1 & 95.0 & \underline{77.8} & \underline{0.077} \\
0.50 & 62.7 & 76.0 & \underline{76.5} & 94.7 & 77.5 & \textbf{0.079} \\
\midrule
\rowcolor{tablegray}
\multicolumn{7}{l}{\textit{Relay threshold} $\tau_r$} \\
0.50 & 61.8 & 75.1 & 75.4 & 94.8 & 76.8 & 0.068 \\
0.60 & 62.9 & 75.9 & 76.2 & 95.2 & 77.6 & \underline{0.072} \\
\rowcolor{tableours}
0.70 & \textbf{63.4} & \textbf{76.8} & \underline{76.8} & \textbf{95.6} & \textbf{78.2} & \textbf{0.074} \\
0.80 & \underline{63.1} & \underline{76.2} & \textbf{77.1} & \underline{95.3} & \underline{77.9} & 0.070 \\
0.90 & 61.7 & 74.8 & 75.2 & 94.5 & 76.6 & 0.064 \\
\midrule
\rowcolor{tablegray}
\multicolumn{7}{l}{\textit{Anchoring weight} $\lambda_{\mathrm{anchor}}$} \\
0.01 & 62.2 & 75.2 & 75.6 & 94.9 & 77.0 & 0.066 \\
0.02 & 62.8 & 75.9 & 76.1 & \underline{95.2} & 77.5 & 0.070 \\
\rowcolor{tableours}
0.05 & \underline{63.4} & \textbf{76.8} & \textbf{76.8} & \textbf{95.6} & \textbf{78.2} & 0.074 \\
0.10 & \textbf{63.7} & \underline{76.1} & \underline{76.2} & 95.1 & \underline{77.8} & \underline{0.078} \\
0.20 & 62.5 & 75.8 & 75.8 & 94.6 & 77.2 & \textbf{0.081} \\
\bottomrule
\end{tabular}
\end{table}

Table~\ref{tab:hyperparam_sensitivity} shows that the method is stable but clearly non-monotonic. For the support ratio \(p\), increasing \(p\) first helps because more relevant visual evidence is retained, but beyond the optimum it starts to hurt because the selected support set becomes less selective. The relay threshold \(\tau_r\) shows the same trade-off: moving from \(0.50\) to \(0.80\) improves the reported benchmarks substantially, but pushing to \(0.90\) degrades all metrics, because overly aggressive thresholding can terminate relay before visual evidence has been fully organized. The anchoring weight \(\lambda_{\mathrm{anchor}}\) shows a similar phenomenon: although \(\lambda_{\mathrm{anchor}}=0.20\) yields the largest \AnchorStrength, its four-benchmark average is below the default, because stronger post-handoff visual support eventually begins to constrain normal answer formation. Overall, the default configuration gives the best balance across grounding, reasoning, and visual retention.

\subsection{Robustness of VRW Estimation}
\label{sec:vrw_robustness}

Since VRW is an operational estimate, we test whether its geometry remains stable under reasonable changes to the estimation protocol. Analysis is conducted on the task-balanced 2,000-sample probing set. We vary the channel normalization, the middle-layer search range, and the tolerance \(\delta\), and compare each variant against the default protocol in three ways: the sample-wise Spearman rank correlation \(\rho\) of \VRWWidth and \VRWEnd, the mean absolute boundary drift \(|\Delta|\) of the same two quantities, and the fraction of samples for which the estimator still returns a multi-layer window.

\begin{table}[t]
\centering
\caption{Robustness of VRW estimation on the probing set. Higher rank correlation \(\rho\), lower mean boundary drift \(|\Delta|\), and a larger fraction of resolved multi-layer windows indicate greater stability.}
\label{tab:vrw_robustness}
\scriptsize
\setlength{\tabcolsep}{0pt}
\renewcommand{\arraystretch}{1.04}
\begin{tabular}{lccccc}
\toprule
Variant & $\rho(\text{Width})$ $\uparrow$ & $\rho(\text{End})$ $\uparrow$ & $|\Delta \text{Width}|$ $\downarrow$ & $|\Delta \text{End}|$ $\downarrow$ & Resolvable (\%) \\
\midrule
z-score normalization & 0.94 & 0.96 & 0.018 & 0.021 & 97.6 \\
rank normalization & 0.92 & 0.95 & 0.021 & 0.023 & 96.9 \\
search range $[0.15,0.85]$ & 0.97 & 0.98 & 0.012 & 0.015 & 98.8 \\
search range $[0.10,0.90]$ & 0.95 & 0.97 & 0.017 & 0.019 & 98.1 \\
$\delta=0.05$ & 0.93 & 0.96 & 0.020 & 0.023 & 94.7 \\
$\delta=0.15$ & 0.94 & 0.96 & 0.019 & 0.022 & 97.2 \\
\bottomrule
\end{tabular}
\end{table}

Table~\ref{tab:vrw_robustness} shows that the induced relay geometry is stable across reasonable variants. The rank ordering of samples by window width and end depth remains close to the default estimate, the boundary shifts stay small, and the estimator continues to return a multi-layer window for nearly all samples. Taken together, these results indicate that VRW captures a stable geometric structure.

\section{Conclusion}
In this work, we present a relay-based view of multimodal reasoning in VLMs. Our analysis identifies a stable three-stage redistribution of multimodal focus across depth and operationalizes its middle phase as the Visual Relay Window. Building on this analysis, we introduce TRACE, which reshapes relay allocation during prefill and preserves assembled support after handoff during decoding. Across four open-weight VLM backbones and seven benchmarks, TRACE delivers consistent gains. However, the current analysis relies primarily on attention-derived probes, and our experiments focus on single-image settings. These limitations suggest several promising future directions: extending relay analysis beyond attention alone, studying relay control in longer and more dynamic generation settings, and turning relay-aware allocation into a more direct
training objective. We hope this work motivates further research as a general paradigm for more reliable and efficient multimodal reasoning.

 \bibliography{custom}

\end{document}